\pdfoutput=1
\newcommand{\CLASSINPUTtoptextmargin}{19mm}
\newcommand{\CLASSINPUTbottomtextmargin}{19mm}
\newcommand{\CLASSINPUTinnersidemargin}{19mm}
\newcommand{\CLASSINPUToutersidemargin}{19mm}
\documentclass[conference]{IEEEtran}
\usepackage{amsmath,amsfonts}
\usepackage{url}
\usepackage{graphicx}
\usepackage{cite}
\usepackage[hidelinks,breaklinks]{hyperref}
\begin{document}

\title{\vspace{6mm}From Simulation to the Real-World: An In-Field 6D Pose Dataset and Baseline for Robotic
Strawberry Harvesting}

\author{
  \IEEEauthorblockN{Woojung Son\IEEEauthorrefmark{1},
    Won Suk Lee\IEEEauthorrefmark{1},
    Zijing Huang\IEEEauthorrefmark{1},
    Daeun Choi\IEEEauthorrefmark{1},
    Catia Silva\IEEEauthorrefmark{2},
    Yu She\IEEEauthorrefmark{3},
    and Yan Gu\IEEEauthorrefmark{4}}\\
  \IEEEauthorblockA{\IEEEauthorrefmark{1}Department of Agricultural and Biological
    Engineering, University of Florida\\
    \{w.son, wslee, zijing.huang, dana.choi\}@ufl.edu}\\
  \IEEEauthorblockA{\IEEEauthorrefmark{2}Department of Electrical and Computer
    Engineering, University of Florida, catiaspsilva@ece.ufl.edu}\\
  \IEEEauthorblockA{\IEEEauthorrefmark{3}Edwardson School of Industrial Engineering,
    Purdue University, yushe@purdue.edu}\\
  \IEEEauthorblockA{\IEEEauthorrefmark{4}School of Mechanical Engineering,
    Purdue University, yangu@purdue.edu}
}

\maketitle
\begin{abstract}
Robotic strawberry harvesting requires precise 6D pose estimation; however, collecting 6D
pose ground truth in
real agricultural fields is inherently challenging. Existing strawberry 6D pose estimation studies have therefore relied mainly on synthetic data, leaving their in-field performance unquantified.
In this work, we obtain ground truth indirectly, by recovering camera poses via PnP,
reconstructing each scene at metric scale, and annotating a single 3D bounding box per
strawberry that is propagated across all frames, yielding, to the best of our knowledge,
the first real-world 6D pose ground-truth dataset of red-stage strawberries collected at
an actual strawberry farm (12{,}040 images). We also introduce a synthetic dataset
rendered in NVIDIA Isaac Sim, featuring scene-level realism
and domain randomization. Despite this improved simulation setup, models trained on synthetic
data alone fail to transfer to in-field images, while introducing a small amount of real data
improves both translation and rotation accuracy across all backbones. Under the monocular
RGB setting evaluated here, rotation is largely recovered once real data is used, and depth
is what limits pose accuracy. These baselines across backbone encoders serve as a reference
for future work.
The real-world dataset is publicly available at \url{https://huggingface.co/datasets/WoojungSon/FieldStraw6D}, and the data-collection pipeline is available at \url{https://github.com/wjson2435/FieldStraw6D-pipeline}.
\end{abstract}


\begin{figure*}[!t]
  \centering
  \includegraphics[width=\textwidth]{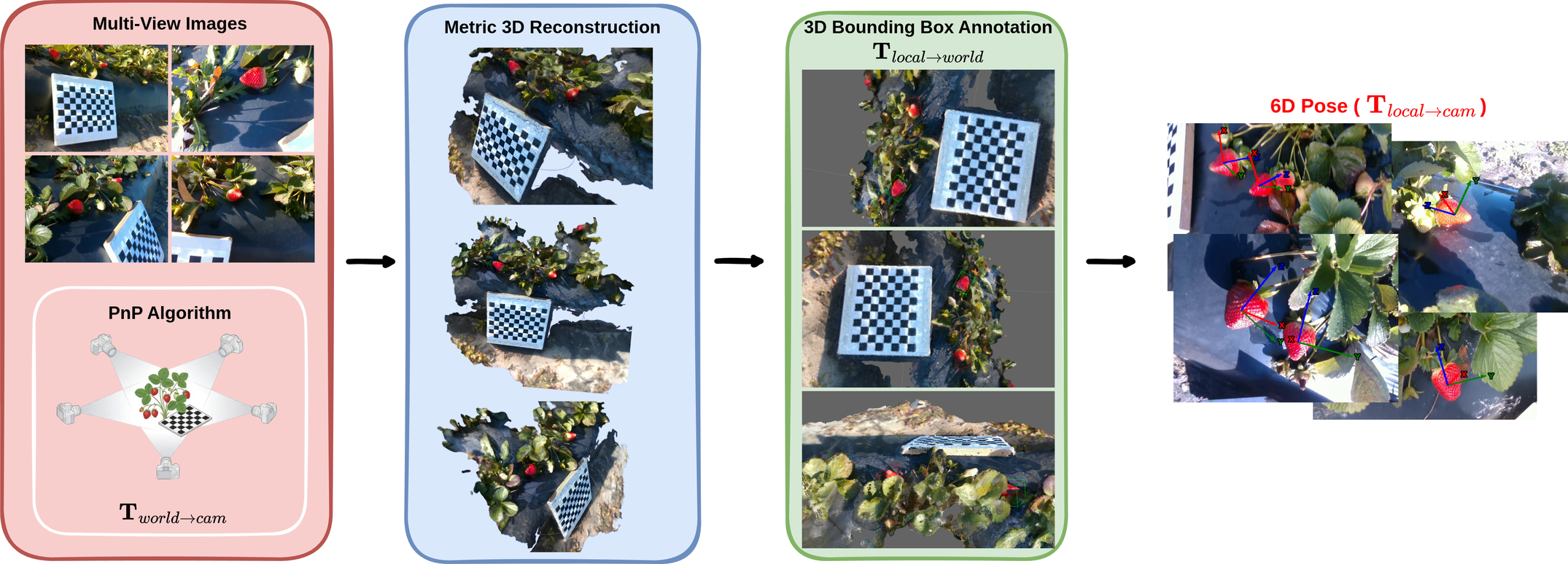}
  \caption{Overview of the real-world dataset construction pipeline.
  A video sequence is recorded with a checkerboard placed near the strawberry
  plant, and PnP is applied to estimate the world-to-camera transformation
  $\mathbf{T}_{world \rightarrow cam}$ for each frame.
  Metric 3D reconstruction is performed using COLMAP, and 3D bounding boxes
  are manually annotated on the resulting point cloud to obtain
  $\mathbf{T}_{local \rightarrow world}$.
  In parallel, 2D bounding boxes are manually annotated on the RGB frames.
  The 6D pose ground truth is finally composed as
  $\mathbf{T}_{local \rightarrow cam} = \mathbf{T}_{world \rightarrow cam}
  \cdot \mathbf{T}_{local \rightarrow world}$.}
  \label{fig:dataset_pipeline}
\end{figure*}

\section{Introduction}

Strawberries are among the most widely consumed fruits in the world, but they are highly
labor-intensive to produce, with harvesting labor accounting for roughly 40\% of total
production costs~\cite{hernandez2023strawberry}, and the industry faces a growing labor
shortage. Fully autonomous strawberry harvesting at commercial scale has yet to be
realized~\cite{defterli2016review}. Among the challenges, damage-free handling of soft,
fragile strawberries requires precise robotic grasping, which depends on knowing not only
fruit position but also orientation, that is, the fruit's full 6D pose. Existing harvesting
systems, however, estimate 3D fruit position only~\cite{xiong2020autonomous,ge2019fruit,ren2024mobile}.

Despite this need, the few methods that do pursue strawberry orientation or 6D pose
estimation~\cite{wagner2021orientation, li2024singleshot, sinha2025strawberry6d} are each
trained and evaluated on their own synthetic dataset. The accuracy they report is therefore
an in-distribution result, measured on the same generator it was trained on, and cannot
predict performance in the field. On real images they can only be inspected qualitatively,
because no means exists of obtaining 6D pose ground truth there, leaving the sim-to-real gap
for this task unmeasured. Measuring this gap is what would make it possible to verify how a
trained model actually performs in the field, and to compare methods on common real data
rather than each on its own simulator.

Such a means has not existed for two reasons. Physically, field-grown plants cannot be
instrumented or marked, and controlled acquisition setups such as turntables are unavailable
in the field. Separately, conventional 6D pose annotation pipelines require an exact CAD
model per object, an assumption strawberries violate given their variable size, shape, and
calyx geometry. Ground truth for agricultural crops has therefore been obtained either in
simulation~\cite{li2024singleshot, sinha2025strawberry6d}, where these obstacles do not
arise, or in controlled laboratory settings~\cite{abdulsalam2023fruity, chatzis2026pear},
where the plant can be instrumented and modeled. Both approaches circumvent these obstacles
by removing the field itself, which is why the obstacles persist wherever the fruit actually
grows.

Because direct in-field measurement is difficult, we obtain ground truth
indirectly. We collect video sequences of
strawberry plants in real fields, recover camera poses via Perspective-n-Point (PnP)
estimation, reconstruct the scene at metric scale, and annotate a 3D bounding box once per
instance in the reconstructed space, propagating it across all frames through the recovered
camera poses. This yields
12,040 annotated images. In parallel, we render a synthetic dataset in NVIDIA Isaac Sim with
scene-level realism and domain randomization, as a control: by improving
simulation fidelity over prior work, we narrow, though do not eliminate, the extent to which
any residual sim-to-real gap can be explained by inadequate rendering. Building on both
datasets, we evaluate three backbone encoders for monocular RGB-only strawberry 6D pose
estimation. Models trained purely on synthetic data transfer poorly to the field, while
adding even a small amount of real data improves both translation and rotation accuracy
across all backbones. Under this monocular RGB setting, rotation is largely recovered once
real data is used, and depth is what limits pose accuracy.

The main contributions of this work are as follows:
\begin{itemize}
\item We present, to the best of our knowledge, the first in-field red-stage strawberry 6D pose ground-truth dataset, containing 12,040 images and 16,037 annotated strawberry instances.
\item We introduce a synthetic dataset rendered in NVIDIA Isaac Sim with
scene-level realism and domain randomization, serving as a control that narrows the extent
to which the measured gap can be attributed to rendering fidelity alone.
\item We provide baseline results across three backbone encoders under synthetic-only, mixed,
and real-only training, quantifying the sim-to-real gap for strawberry 6D pose estimation
in the field.
\end{itemize}

\section{Related Work}

Prior robotic strawberry harvesting systems achieve fruit detection and localization using
RGB-D 3D centroid estimation for pick-and-place operations~\cite{xiong2020autonomous,ge2019fruit,ren2024mobile},
without recovering fruit orientation. Wagner et al.~\cite{wagner2021orientation} estimate
strawberry orientation but not full 6D pose, and Li and Kasaei~\cite{li2024singleshot} and
Sinha et al.~\cite{sinha2025strawberry6d} extend this to full 6D pose estimation. All three
train purely in simulation and report quantitative accuracy only on synthetic test data;
on real images they are restricted to qualitative inspection, as no in-field 6D pose
ground truth exists to score against.

While substantial effort has gone into agricultural perception datasets, few provide 6D pose
ground truth for agricultural produce. Abdulsalam et al.~\cite{abdulsalam2023fruity} and
Chatzis et al.~\cite{chatzis2026pear} introduce multi-category produce datasets with 6D pose
ground truth, collected in controlled laboratory settings, including through a calibrated
robotic manipulator~\cite{chatzis2026pear}; neither includes strawberries. Chatzis et al.
further note that instance-level 6D pose approaches require exact 3D CAD models, a
limitation for agricultural produce with high intra-class shape variability.

Synthetic data is widely used in agricultural robotics to reduce annotation cost, but a
domain gap between simulated and real environments persists even for simpler tasks such as
object detection: Hutter-Mironovová~\cite{hutter2026simtoreal} reports that fruit detectors
trained exclusively on synthetic images fall well short of their real-trained counterparts, and
that while adding real images recovers most of this gap in-domain, hybrid models still
degrade once the background changes. Because 6D
pose estimation demands finer-grained geometric understanding than detection, this gap may be
more severe for pose estimation, yet for agricultural produce it remains unmeasured
in the field.

\section{Method}
We first describe the construction of the real-world and synthetic datasets
(Sec.~\ref{sec:dataset}), then the baseline 6D pose estimation architecture trained on them
(Sec.~\ref{sec:baseline}).

\subsection{Dataset}
\label{sec:dataset}
  We collect two complementary datasets for strawberry 6D pose estimation:
  a real-world dataset of 12{,}040 images captured in an actual agricultural
  field, and a synthetic dataset rendered in NVIDIA Isaac Sim. For both datasets, the strawberry's local coordinate frame is defined
  with the $+z$-axis pointing toward the stem. The $x$- and $y$-axes are not
  anchored to a specific anatomical feature; they follow the orientation of the
  annotated (real-world) or modeled (synthetic) bounding box, since strawberries
  are approximately rotationally symmetric about the stem axis and lack a
  consistent visual landmark for a canonical azimuthal reference. We revisit
  this in Sec.~\ref{sec:eval_metrics} with a symmetry-aware stem-axis error
  that is invariant to this azimuthal ambiguity. Dataset details are
  summarized in Table~\ref{tab:dataset}.

\begin{table}[t]
  \centering
  \caption{Overview of the real-world and synthetic datasets.
         Instances denote the total number of annotated strawberries
         across all images.}
  \label{tab:dataset}
\begin{tabular}{lcc}
  \hline
  & Real-World & Synthetic \\
  \hline
  Images      & 12,040  & 35,118  \\
  Instances   & 16,037  & 127,910 \\
  Resolution  & \multicolumn{2}{c}{$640 \times 480\,\text{pixels}$} \\
  Sensor / Renderer & RealSense D435i & Isaac Sim \\
  Environment & Strawberry Farm & Simulated Strawberry Farm \\
  Annotation  & \multicolumn{2}{c}{6D pose + 2D bbox} \\
  \hline
\end{tabular}
\end{table}

 \noindent\textbf{Real-World Dataset.}
  The construction pipeline is illustrated in Fig.~\ref{fig:dataset_pipeline}.
  An Intel RealSense D435i camera is used for all data collection,
  capturing frames at $640 \times 480\,\text{pixels}$ resolution.
  Since robotic harvesting targets only ripe fruit, data collection is
  restricted to red-stage strawberries.
  We first calibrate the camera to obtain the intrinsic matrix $\mathbf{K}$,
  required by the PnP algorithm in the subsequent step.

  A checkerboard ($11 \times 8$ squares, A4, 25\,mm square size) is placed near
  the strawberry plant, and a video sequence is recorded from varying distances,
  within the limits imposed by the surrounding plants and bed layout, to cover
  viewpoints relevant to robotic manipulation, increase the diversity of 6D pose
  annotations, and improve the success rate of 3D reconstruction.
  Frames are extracted from the recorded video, and for frames in which the
  checkerboard is visible, we apply the PnP algorithm~\cite{lepetit2009epnp}
  to compute the world-to-camera transformation $\mathbf{T}_{world \rightarrow cam}$.
  We represent all transformations as $4 \times 4$ homogeneous matrices:
  \begin{equation}
    \mathbf{T} = \begin{bmatrix} \mathbf{R} & \mathbf{t} \\ \mathbf{0}^\top & 1
    \end{bmatrix} \in SE(3), \quad \mathbf{R} \in SO(3), \quad \mathbf{t} \in \mathbb{R}^3
  \end{equation}

\begin{figure*}[t]
  \centering
  \includegraphics[width=\textwidth]{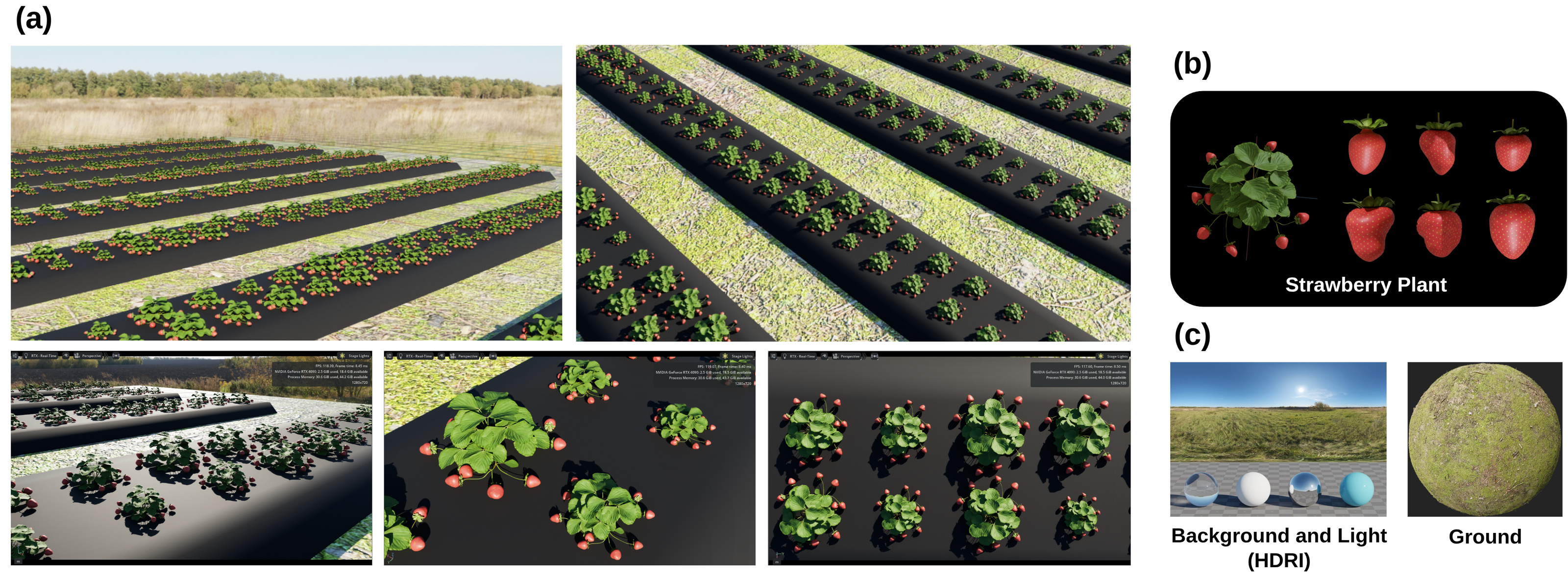}
  \caption{Synthetic environment in NVIDIA Isaac Sim.
  \textbf{(a)} Rendered strawberry farm scenes.
  \textbf{(b)} Strawberry plant models with geometry variation.
  \textbf{(c)} Background and environment lighting from a high dynamic range
  image (HDRI), and ground material.}
  \label{fig:synthetic}
\end{figure*}

  We perform sparse reconstruction with COLMAP~\cite{schoenberger2016sfm,
  schoenberger2016mvs} on all extracted frames, aligning the reconstructed model
  to the world coordinate system using the per-frame $\mathbf{T}_{world
  \rightarrow cam}$ as a reference pose to recover metric scale.
  Because COLMAP registers all frames of a sequence into a single
  reconstruction, the checkerboard needs only to be visible in a subset of frames:
  those frames anchor the reconstruction to the world coordinate system, and the
  multi-view constraints between cameras then place every remaining frame in the
  same coordinate system, yielding $\mathbf{T}_{world \rightarrow cam}$ for all
  frames.

  Dense reconstruction is then performed via PatchMatch stereo and stereo fusion
  to produce a metric point cloud. A sequence is discarded only when COLMAP fails
  to produce a reconstruction; no reprojection-error threshold is applied. We
  report the reprojection errors of the retained sequences as a measure of
  reconstruction quality in Sec.~\ref{sec:dataset_analysis}.

  We manually annotate a 3D bounding box on the reconstructed metric point cloud for each visible strawberry instance, specifying its position, orientation, and dimensions.
  Each annotated box provides the object's position and orientation in the world
  coordinate system as $\mathbf{T}_{local \rightarrow world}$.

  The 6D pose of each object with respect to the camera is then obtained by:
  \begin{equation}
    \mathbf{T}_{local \rightarrow cam} =
    \mathbf{T}_{world \rightarrow cam} \cdot \mathbf{T}_{local \rightarrow world}
    \label{eq:pose}
  \end{equation}

  To align with the OpenGL coordinate convention adopted by NVIDIA Isaac Sim, the resulting
  poses are converted from OpenCV by applying a $180^\circ$ rotation about the $x$-axis:
  \begin{equation}
    \mathbf{T}_{\text{OpenGL}} = \mathbf{M} \cdot \mathbf{T}_{\text{OpenCV}},
    \quad
    \mathbf{M} = \begin{bmatrix}
      1 &  0 &  0 & 0 \\
      0 & -1 &  0 & 0 \\
      0 &  0 & -1 & 0 \\
      0 &  0 &  0 & 1
    \end{bmatrix}
  \end{equation}

  Finally, 2D bounding boxes are manually annotated for instances belonging to
  successfully reconstructed sequences, i.e., sequences not discarded due to COLMAP
  reconstruction failure (Sec.~\ref{sec:dataset_analysis}).
  The resulting real-world dataset comprises 12,040 images and 16,037 annotated strawberry instances, each with a 6D pose and a 2D bounding box.

 \noindent\textbf{Synthetic Dataset.}
  Prior synthetic strawberry datasets model plants inside a general-purpose
  robotics simulator~\cite{li2024singleshot} or place fruit among
  non-agricultural distractor objects~\cite{sinha2025strawberry6d}, without
  reproducing the surrounding field environment.
  Our dataset instead models a full strawberry farm scene, with realistic
  High Dynamic Range Image (HDRI) environment lighting, a textured ground
  material, and plant-level geometry variation.
  The HDRI is a panoramic photograph of a real outdoor scene that both
  illuminates the virtual scene and serves as its background.
  We construct a strawberry farm environment in NVIDIA Isaac Sim using a
  red-stage strawberry plant model,
  a CC0-licensed ground material from AmbientCG~\cite{ambientcg},
  and a farm field HDRI from Poly Haven~\cite{polyhaven}.
  To ensure domain alignment, the synthetic camera is configured to match
  the Intel RealSense D435i used in real-world collection, including identical
  image resolution ($640 \times 480\,\text{pixels}$) and camera intrinsics.

  The camera is positioned on a hemisphere centered on the strawberry plant,
  with the distance sampled uniformly in $[0.2, 1.0]$\,m and the viewing direction
  constrained to face the plant at all times.
  To reduce positional bias in the synthetic images,
  a random offset is applied to the look-at target point, so that the camera
  points toward a location near, but not exactly at, the strawberry center,
  causing the strawberry to appear at varying 2D positions within the image
  rather than always at the center.
  Domain randomization is applied over lighting conditions (intensity and
  direction) and strawberry plant geometry (size and shape),
  with a randomized seed for each scene to maximize dataset diversity.
  For each frame, we collect the RGB image, 2D bounding box,
  3D bounding box (which provides $\mathbf{T}_{local \rightarrow world}$),
  and camera extrinsics (which provide $\mathbf{T}_{world \rightarrow cam}$),
  all provided directly by the simulator without manual annotation.
  The 6D pose ground truth is then derived following the same formulation
  as the real-world dataset (Eq.~\eqref{eq:pose}).
  Licensing rights to the strawberry plant model are held by one of the authors
  and do not permit open redistribution, so the model and the rendered synthetic
  dataset are available upon request.

\subsection{Baseline Architecture}
\label{sec:baseline}

\begin{figure}[t]
  \centering
  \includegraphics[width=\columnwidth]{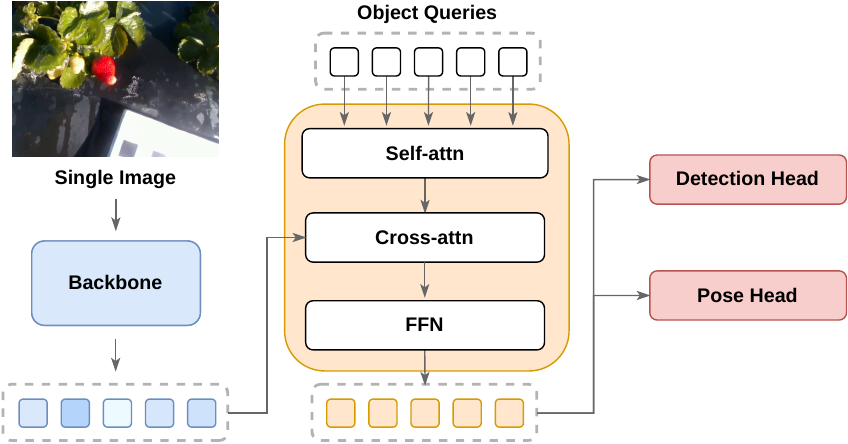}
  \caption{Overview of the baseline architecture. Given a single RGB image,
  a backbone encoder extracts patch-level features, which are then processed
  by a transformer decoder with learnable object queries to jointly
  predict 2D detections and 6D poses via dedicated detection and
  pose heads.}
  \label{fig:baseline}
\end{figure}

The overall architecture is shown in Fig.~\ref{fig:baseline}.
We evaluate three backbone encoders representing different architectural and pretraining settings: ResNet-101~\cite{he2016resnet} as a CNN baseline, ViT-B/16~\cite{dosovitskiy2021vit} as a supervised transformer baseline, and DINOv2-B~\cite{oquab2024dinov2} as a self-supervised vision foundation model.

All three backbones are evaluated within a shared architecture, where spatial features extracted by the backbone encoder are fed into a DETR-style~\cite{carion2020detr} transformer decoder with learned object queries. The decoder jointly
predicts a 2D bounding box, rotation represented via the continuous 6D
representation of Zhou et al.~\cite{zhou2019continuity}, and translation
decoupled into in-plane ($x$, $y$) and depth ($z$) components.

\noindent\textbf{Loss Function.}
Object queries are matched to ground-truth annotations via Hungarian matching
with cost:
\begin{equation}
    \mathcal{C} = \lambda_{\text{bbox}}\mathcal{C}_{\text{bbox}} +
\lambda_{\text{giou}}\mathcal{C}_{\text{giou}}
\end{equation}
where $\mathcal{C}_{\text{bbox}}$ is the L1 distance between predicted and
ground-truth boxes and $\mathcal{C}_{\text{giou}}$ is the negative GIoU.
Since only a single class is present, the classification term is omitted from
the matching cost.
Rotation and translation losses are computed only over matched pairs.
The total training loss is:
\begin{equation}
    \mathcal{L} = \mathcal{L}_{\text{det}} +
    \alpha(t)\!\left(\lambda_{\text{rot}}\mathcal{L}_{\text{rot}}
    + \lambda_{\text{xy}}\mathcal{L}_{\text{xy}} + \lambda_{z}\mathcal{L}_{z}\right)
\end{equation}
where $\mathcal{L}_{\text{det}} = \lambda_{\text{cls}}\mathcal{L}_{\text{cls}} +
\lambda_{\text{bbox}}\mathcal{L}_{\text{bbox}} + \lambda_{\text{giou}}\mathcal{L}_{\text{giou}}$,
$\mathcal{L}_{\text{cls}}$ is the object-vs-background classification loss,
$\mathcal{L}_{\text{rot}}$ is the geodesic loss on rotation matrices,
and $\mathcal{L}_{\text{xy}}$, $\mathcal{L}_{z}$ are SmoothL1 losses on
in-plane and depth translation, respectively.

Since object queries must first learn to reliably localize strawberries
before pose predictions become meaningful, pose losses are activated
gradually via a warmup factor $\alpha(t) = \min(t / T_{\text{warmup}}, 1)$,
where $t$ is the current epoch and $T_{\text{warmup}}$ is the warmup duration.

\section{Experiments}

\subsection{Experimental Setup}

\noindent\textbf{Hardware.}
All models are trained on a single NVIDIA B200 GPU of HiPerGator, the
University of Florida's high-performance computing cluster.

\noindent\textbf{Dataset Split.}
The real-world dataset is partitioned at the frame (image) level into training
(10,040 images), validation (1,000 images), and test (1,000 images) sets.
Validation and test sets consist exclusively of real-world images, and their
indices are fixed prior to all experiments. The training set
size is fixed at 10,040 images, with the proportion of real-world data
varied from 0\% (synthetic-only) to 100\% (real-only), with the remainder
drawn from the synthetic dataset. We denote the mixed settings as
\textit{R1}--\textit{R9}, corresponding to 10\%--90\% real-world training
data, respectively.

\noindent\textbf{Training Details.}
All models are trained using the AdamW optimizer with a learning rate of
$1 \times 10^{-4}$ and weight decay of $1 \times 10^{-4}$. A StepLR
scheduler reduces the learning rate every 30 epochs, and gradient clipping
is applied with a maximum norm of 0.1. All backbones are trained for 50 epochs, using a batch size of 256.
DINOv2-B is kept frozen, so only the decoder parameters are optimized, whereas
ResNet-101 and ViT-B/16 are fine-tuned with the backbone updated at $0.1\times$
the base learning rate to preserve pretrained representations. This asymmetry
should be kept in mind when comparing backbones: DINOv2-B achieves the strongest
results in our experiments despite receiving no backbone adaptation at all.
To stabilize early training, pose losses are activated via a linear warmup
with $T_{\text{warmup}} = 20$ epochs, i.e., $\alpha(t) = \min(t/T_{\text{warmup}}, 1)$.

\begin{figure*}[t]
  \centering
  \includegraphics[width=\textwidth]{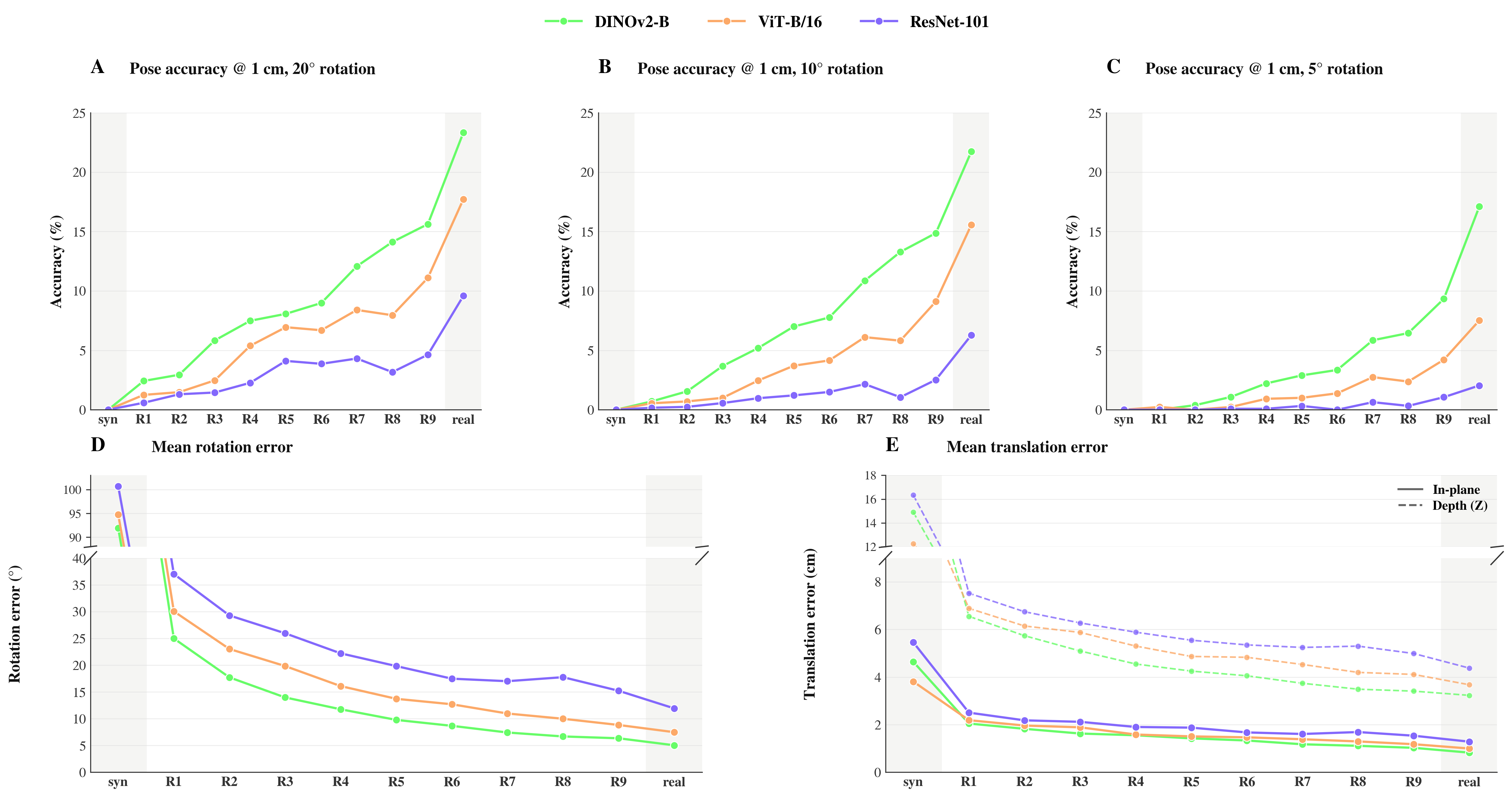}
  \caption{Baseline results across backbone architectures and training
  configurations. \textit{Syn} denotes synthetic-only training;
  \textit{R1--R9} denote mixed training sets with 10\%--90\% real-world data;
  \textit{Real} denotes real-only training. Gray shaded regions indicate the
  synthetic-only and real-only training configurations.
  \textbf{(A--C)} Pose accuracy at the stricter, grasp-relevant
  $\tau_t = 1\,\text{cm}$ translation threshold, with rotation thresholds of
  $20^\circ$, $10^\circ$, and $5^\circ$, respectively.
  \textbf{(D)} Mean rotation error. \textbf{(E)} Mean translation error
  decomposed into in-plane ($\sqrt{\epsilon_x^2+\epsilon_y^2}$) and depth
  ($\epsilon_z$) components. Panels D and E use a broken $y$-axis, since the
  synthetic-only value lies far outside the range of the real-data mixtures.}
  \label{fig:results_summary}
\end{figure*}

\noindent\textbf{Loss Coefficients.}
Following DETR~\cite{carion2020detr}, detection losses use coefficients of
1.0 (classification), 5.0 (bounding box L1), and 2.0 (GIoU).

Pose losses are weighted at 10.0 for rotation and in-plane translation,
and 15.0 for depth translation, reflecting the inherently greater difficulty
of monocular depth estimation.

\subsection{Evaluation Metrics}
\label{sec:eval_metrics}

Since strawberries are organic objects with substantial intra-class shape variation, a single mesh model cannot represent the geometry of individual real-world instances, making ADD (Average Distance of Model Points) and ADD-S (Average Distance of Model Points for Symmetric objects) unsuitable without instance-specific mesh models.
We therefore adopt the following metrics.

\begin{figure}[t]
  \centering
  \includegraphics[width=\columnwidth]{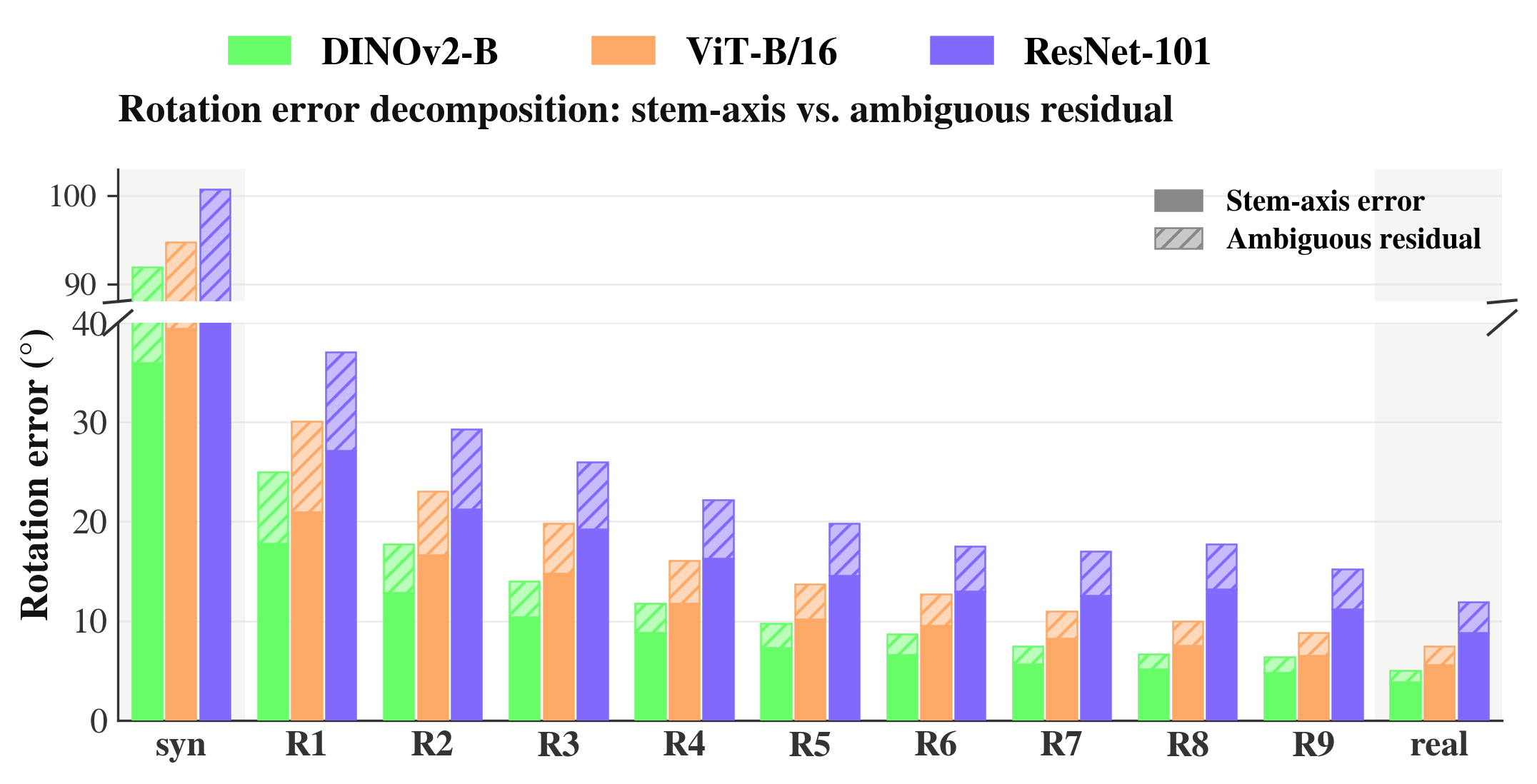}
  \caption{Full rotation error versus the symmetry-aware stem-axis error
  (Sec.~\ref{sec:eval_metrics}), which isolates the predicted stem-pointing
  direction from rotation about that axis. \textit{Syn}, \textit{R1--R9},
  \textit{Real}, and gray shaded regions as in Fig.~\ref{fig:results_summary}.
  The plot uses a broken $y$-axis, as in Fig.~\ref{fig:results_summary}D--E.}
  \label{fig:stem_axis}
\end{figure}
\noindent\textbf{Mean Rotation Error} ($^\circ$) reports the average geodesic error between predicted and ground-truth rotations.

\noindent\textbf{Stem-Axis Error} ($^\circ$) reports the angle between the
predicted and ground-truth stem-direction ($+z$) vectors, ignoring rotation
about that axis. Since the $x$/$y$ azimuthal reference is not anatomically
controlled (Sec.~\ref{sec:dataset}) and strawberries are approximately rotationally
symmetric about the stem, this symmetry-aware metric isolates the component of
orientation (which way the stem points) that is both well-defined in the
annotations and directly relevant to a grasp approach, complementing the full
rotation error above.
Our model regresses the full 3D rotation throughout; the stem direction
reported here is one column of the predicted rotation matrix, so no
axis-only detector is trained. Retaining the full rotation keeps the
baseline comparable to standard 6D pose formulations and preserves the
azimuthal component for grasp planners that may exploit it, for instance to
steer a gripper around the calyx. It also makes the two quantities
separable, which we use in Sec.~\ref{sec:results} to examine how the rotation
error is divided between the stem direction and the azimuthal component across
training configurations.

\noindent\textbf{Mean Translation Error} (cm) reports translation error in centimeters and is further decomposed into in-plane ($\sqrt{\epsilon_x^2 + \epsilon_y^2}$) and depth components ($\epsilon_z$).

\noindent\textbf{Pose Accuracy} is the fraction of predictions
for which both the translation error and rotation error fall below
their respective thresholds $\tau_t$ and $\tau_r$.
As a concrete point of reference, a ripe commercial strawberry is typically
about $3$--$4\,\text{cm}$ in diameter; a $1\,\text{cm}$ error therefore
corresponds to roughly a quarter to a third of the fruit's diameter, large
enough to plausibly affect grasp alignment, motivating our choice of
$1\,\text{cm}$ as a deliberately strict, task-scale-grounded threshold
rather than an arbitrary one.
We evaluate translation accuracy at this stricter, grasp-relevant
$\tau_t = 1\,\text{cm}$ threshold throughout, chosen to make the gap between
estimator capability and task requirement explicit rather than leaving it
implicit.
We vary $\tau_r \in \{5^\circ, 10^\circ, 20^\circ\}$ to assess
rotational accuracy at multiple strictness levels.

\subsection{Results}
\label{sec:results}

\begin{figure*}[t]
  \centering
  \includegraphics[width=\textwidth]{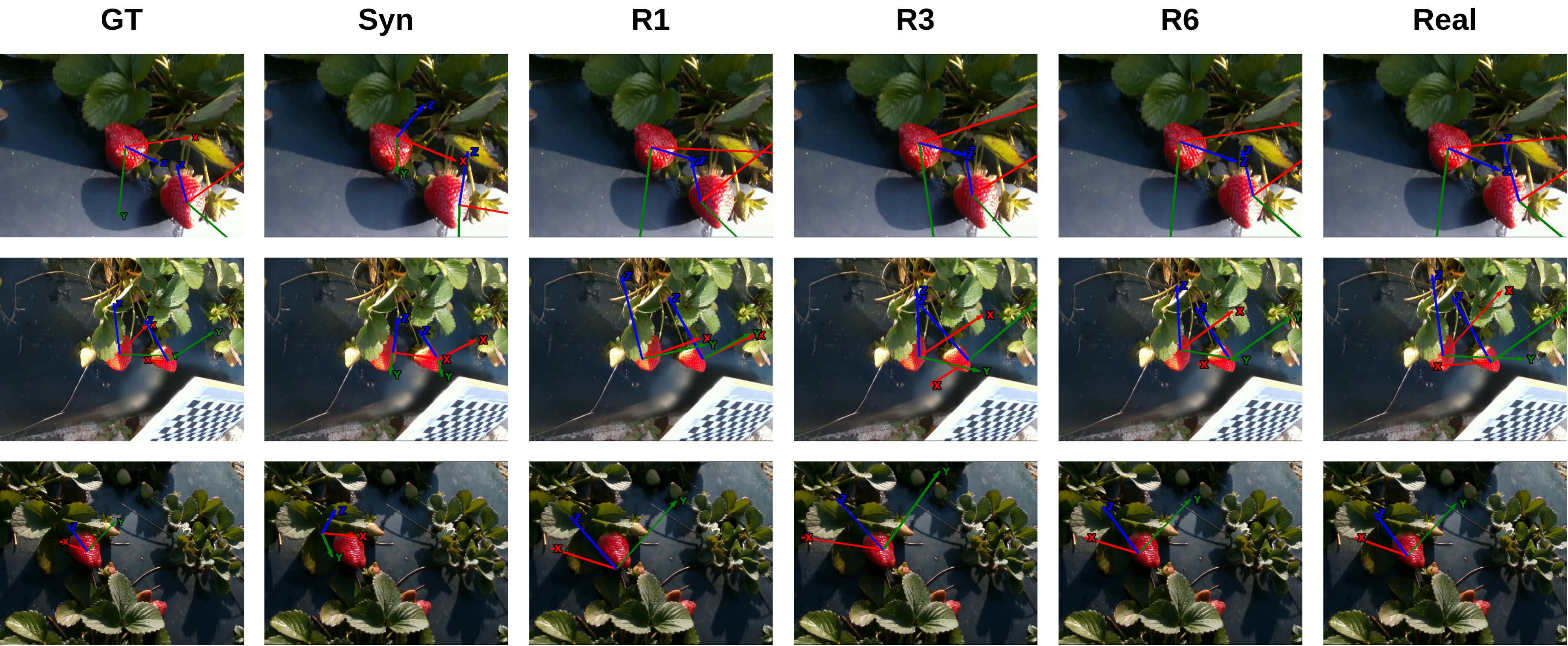}
  \caption{Qualitative 6D pose estimation results across training configurations
  (DINOv2-B backbone).
  Columns correspond to: ground truth (GT), synthetic-only training (\textit{Syn}),
  and models trained with increasing amounts of real data (\textit{R1}, \textit{R3}, \textit{R6}, \textit{Real}).
  Coordinate axes are color-coded: X (red), Y (green), Z (blue).
  \textit{Syn}-only training yields poses far from GT across all scenes.
  Adding even a single real-data increment (\textit{R1}) substantially reduces error, and predictions become progressively closer to GT as more real data is added.}
  \label{fig:qualitative}
\end{figure*}

\noindent\textbf{Synthetic-only training yields poor real-world transfer.}
Models trained exclusively on synthetic data achieve 0.0\% pose accuracy on real
agricultural field images across all evaluated backbones and thresholds.
Mean rotation errors exceed $90^\circ$ for all backbones
(DINOv2-B: $91.90^\circ$, ViT-B/16: $94.72^\circ$, ResNet-101: $100.68^\circ$),
consistent with the synthetic-only values in Fig.~\ref{fig:results_summary}D.
This demonstrates that even synthetic data with scene-level realism is insufficient for
strawberry 6D pose estimation on an actual farm.

\noindent\textbf{A small fraction of real data yields most of the improvement.}
Introducing just $10\%$ real data (\textit{R1}) causes a dramatic performance jump
across all backbones.
DINOv2-B's mean rotation error drops from $91.90^\circ$ to $25.00^\circ$, a
$73\%$ reduction from a single increment (Fig.~\ref{fig:results_summary}D).
Performance continues to improve as more real data is added, with gains
largest at \textit{R1} and diminishing progressively thereafter; no single saturation
point is evident across all backbones.

\noindent\textbf{Depth error is the dominant translation bottleneck.}
Depth translation error consistently exceeds in-plane error across all
conditions, reflecting the fundamental depth ambiguity of monocular RGB-only 6D pose
estimation.
Under real-only training, DINOv2-B achieves an in-plane error of
$0.83\,\text{cm}$ but a depth error of $3.23\,\text{cm}$, roughly $4\times$
larger (Fig.~\ref{fig:results_summary}E) and already spanning roughly $80\%$ of
the fruit's width.
Notably, in-plane error is already manageable at \textit{R1} (${\sim}2\,\text{cm}$)
and changes little thereafter, whereas depth error decreases more gradually
and remains the primary bottleneck throughout.
Decomposing the accuracy criterion confirms which term binds. Under real-only
training, $92.5\%$ of DINOv2-B's predictions fall within the $10^\circ$
rotation threshold, but only $23.3\%$ fall within the $1\,\text{cm}$
translation threshold, and the joint accuracy of $21.7\%$ is nearly identical
to the latter; the same ordering holds for ViT-B/16 ($81.7\%$ versus $17.7\%$,
joint $15.6\%$) and ResNet-101 ($52.8\%$ versus $10.8\%$, joint $6.3\%$).
Translation, and depth within it, is therefore what limits pose accuracy at
this threshold, while rotation is rarely the failing criterion.
Closing this gap likely requires additional
depth cues (stereo, RGB-D, or multi-view geometry) beyond monocular RGB, which
we leave for future work.
\begin{figure}[t]
  \centering
  \includegraphics[width=\columnwidth]{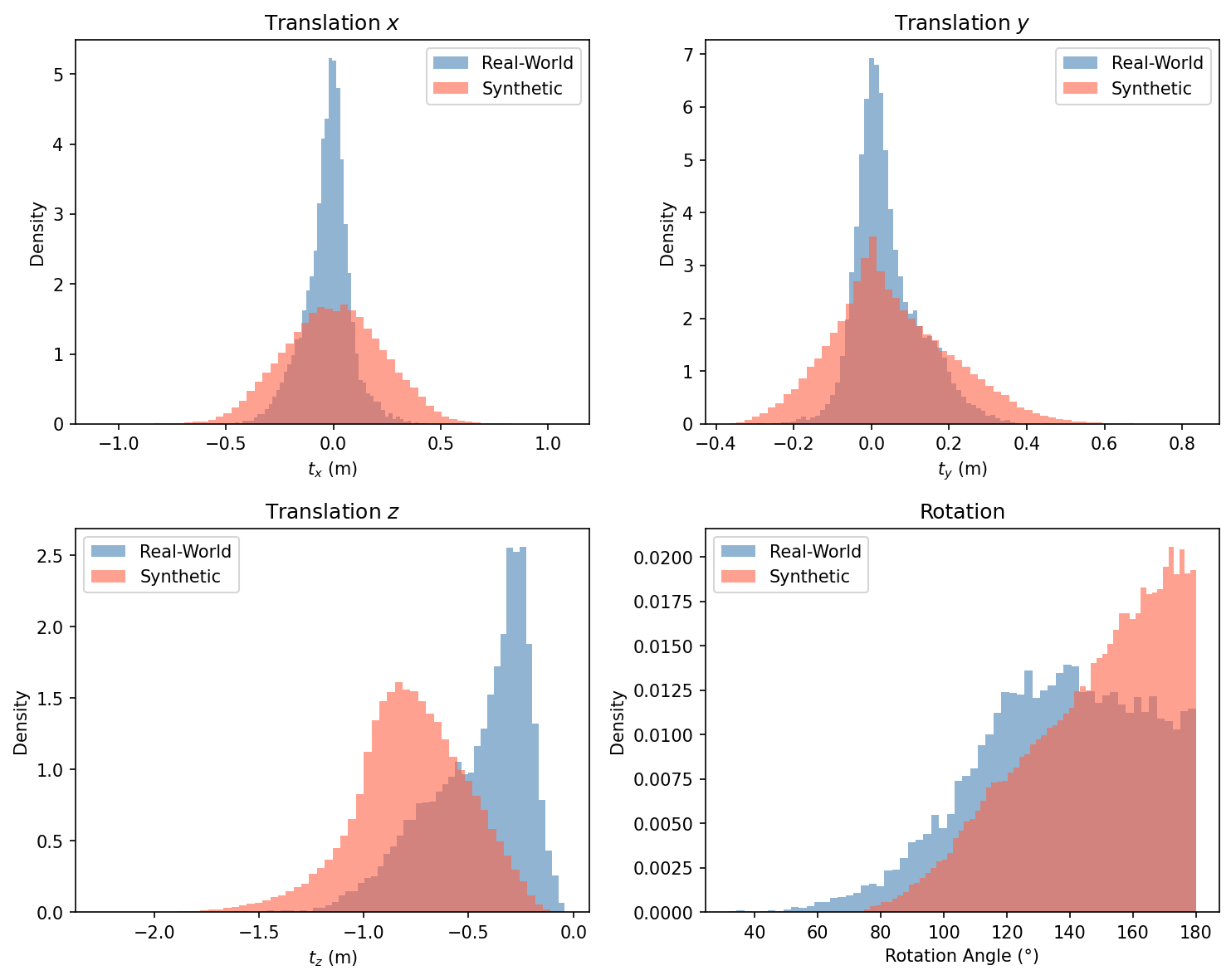}
  \caption{Pose distributions of the real-world and synthetic datasets, including in-plane translations ($t_x$, $t_y$),
  depth translation ($t_z$), and rotation angle, defined as the geodesic magnitude of
  $\mathbf{R}_{local \rightarrow cam}$. The distributions reveal domain shifts in translation
  spread, depth range, and orientation coverage.}
  \label{fig:dataset_dist}
\end{figure}

\noindent\textbf{The azimuthal share of rotation error depends strongly on
training configuration.}
The stem-axis error is consistently lower than the full rotation error across
every backbone and training configuration (Fig.~\ref{fig:stem_axis}), but the
size of the gap differs sharply between conditions. Under real-only training,
it is small in absolute terms: DINOv2-B's full rotation error of $5.04^\circ$
falls to $3.87^\circ$ when azimuthal rotation about the stem is excluded, a
difference of $1.17^\circ$ ($23\%$), with ViT-B/16 and ResNet-101 showing
comparable relative reductions of $26\%$ each ($1.93^\circ$ and $3.10^\circ$).
Under synthetic-only training the gap is far larger:
DINOv2-B's full rotation error of $91.90^\circ$ falls to $35.98^\circ$
($61\%$ lower), with ViT-B/16 showing a similarly large $58\%$ reduction and
ResNet-101 a $52\%$ reduction. Under real-only training the two metrics
nearly coincide, so the rotation accuracy reported above is not an artifact
of azimuthal ambiguity. The large azimuthal share under
synthetic-only training should be read with care: at rotation errors near
$90^\circ$, the orientation is poorly recovered in any case, and a large
azimuthal component is expected simply as a consequence.

\noindent\textbf{Qualitative Analysis.}
Fig.~\ref{fig:qualitative} visualizes predicted poses across training
configurations (DINOv2-B backbone).
Synthetic-only predictions exhibit large axis misalignment across all
scenes, while progressive convergence toward GT is visually apparent
as real data increases.

\subsection{Dataset Analysis}
\label{sec:dataset_analysis}

\noindent\textbf{Ground-Truth Accuracy.}
The accuracy of the real-world ground-truth poses is affected by three
quantifiable sources of error: camera calibration, PnP estimation, and COLMAP
reconstruction, as well as manual 3D bounding box annotation, which we do not
quantify.
Camera calibration achieves an RMS reprojection error of $0.21\,\text{pixels}$.
PnP estimation yields a median reprojection error of $0.50\,\text{pixels}$
across 7{,}156 frames (mean $1.17\,\text{pixels}$, right-skewed due to extreme viewpoints).
COLMAP reports a mean reprojection error of $0.96\,\text{pixels}$ across
99 valid sequences; 21 sequences were discarded due to reconstruction failure.
Of the three quantified stages, COLMAP reconstruction shows the largest
reprojection error; because annotation error is not quantified, we do not
rank the sources overall.

\noindent\textbf{Pose Distributions.} Fig.~\ref{fig:dataset_dist} shows the pose
distributions of both datasets.
For in-plane translation ($t_x$, $t_y$), both distributions share a similar
mean near zero, but differ substantially in spread: real-world instances
exhibit a much narrower concentration, reflecting the constrained camera
positioning during field collection, whereas synthetic instances are spread
over a significantly wider range due to the unconstrained hemisphere sampling.
The depth component $t_z$ reveals a clear distributional shift between domains.
Under the OpenGL convention adopted here (Sec.~\ref{sec:dataset}), the camera looks along
$-z$, so $t_z$ is negative, and we report distances as $|t_z|$.
Real-world strawberries are captured at closer range, roughly $0.3$\,m from the
camera, consistent with the actual robotic manipulation workspace, while
synthetic instances are centered farther away at roughly $0.7$\,m, a consequence of
sampling the synthetic camera distance uniformly over $[0.2, 1.0]$\,m
(Sec.~\ref{sec:dataset}) rather than matching the distribution observed in the field.
Rotation distributions also differ. We summarize each pose by the geodesic magnitude of
$\mathbf{R}_{local \rightarrow cam}$, i.e., the angle of its axis--angle representation, so
larger values indicate fruit orientations further from alignment with the camera frame.
Synthetic data concentrates more heavily toward $160^\circ$--$180^\circ$ than our field data,
over-representing orientations far from alignment with the camera frame.
These distributional mismatches in in-plane translation spread, mean depth, and rotation
plausibly contribute to the sim-to-real gap quantified above, although we did not run a
controlled experiment isolating their effect and therefore cannot attribute the gap to them.

\section{Conclusion}
We presented, to the best of our knowledge, the first real-world 6D pose ground-truth dataset of red-stage
strawberries collected at an actual strawberry farm (12,040 images), alongside a synthetic
dataset rendered in NVIDIA Isaac Sim with scene-level realism
and domain randomization.
Baseline experiments across backbone encoders, under a monocular RGB setting, reveal that a
substantial sim-to-real gap persists despite this improved simulation setup: synthetic-only
training fails to transfer, and while a small amount of real data substantially improves
rotation accuracy, depth error remains the limiting factor throughout. Closing this gap through simulation alone would
require going beyond the scene-level realism and domain randomization explored here; therefore, real
in-field data remains necessary for both training and evaluating strawberry 6D
pose estimation.

Several limitations remain: the dataset comes from a single farm and camera; annotation
noise is uncharacterized, as each point cloud was labeled by one annotator; and the
synthetic data would benefit from a more realistic strawberry plant model and broader
domain randomization over backgrounds, ground materials, and lighting.

\section*{Acknowledgment}
We thank Jessie Lin for assistance with annotation. This work is supported by the Engineering for Agricultural
Production and Processing, project award
no.~\mbox{2024-67021-42878}, from the U.S. Department of Agriculture's
National Institute of Food and Agriculture.

  \bibliographystyle{IEEEtran}
  \bibliography{references}

\end{document}